\documentclass[conference]{IEEEtran}

\usepackage[utf8]{inputenc}
\usepackage[T1]{fontenc}
\usepackage{cite}
\usepackage{amsmath,amssymb,amsfonts}
\usepackage{graphicx}
\usepackage{booktabs}
\usepackage{textcomp}
\usepackage{xcolor}
\usepackage{array}
\usepackage{colortbl}
\usepackage{url}
\usepackage{multirow}
\usepackage[colorlinks=true,linkcolor=black,citecolor=black,urlcolor=blue]{hyperref}

\definecolor{oursorange}{RGB}{255,243,230}

\newcommand{\obs}{$\bullet$}
\newcommand{\miss}{{\scriptsize$\times$}}

\newcommand{\modelhdr}[2]{%
  \begin{tabular}[c]{@{}c@{}}\textbf{#1}\\[-1.5pt]{\tiny #2}\end{tabular}}

\newcommand{\auc}[2]{#1{\tiny$\pm$#2}}
\newcommand{\aucbest}[2]{\textbf{#1}{\tiny$\pm$#2}}
\newcommand{\aucsecond}[2]{\underline{#1}{\tiny$\pm$#2}}
\newlength{\modw}
\newcolumntype{M}{>{\centering\arraybackslash}p{\dimexpr\modw/4-2\tabcolsep\relax}}

\begin{document}

\title{CoPoE: Multimodal Fusion via Decomposable Disease-Coordinate Product-of-Experts for Missing-Modality Alzheimer's Diagnosis}

\author{
\IEEEauthorblockN{Chihun An}
\IEEEauthorblockA{
\textit{Department of Artificial Intelligence} \\
\textit{Hanyang University} \\
Seoul, Republic of Korea \\
chihun@hanyang.ac.kr
}
\and
\IEEEauthorblockN{Ikbeom Jang}
\IEEEauthorblockA{
\textit{Division of Computer Engineering} \\
\textit{Hankuk University of Foreign Studies} \\
Yongin, Republic of Korea \\
ijang2@hufs.ac.kr
}
}

\maketitle

\begin{abstract}
Multimodal Alzheimer's disease (AD) diagnosis benefits from integrating heterogeneous clinical, imaging, genomic, and biomarker evidence, but clinical cohorts frequently suffer from irregular modality missingness. Existing fusion methods often synthesize absent inputs, risking the introduction of artificial surrogates, or pool available signals into uninterpretable latent spaces. We present CoPoE (Disease-Coordinate Product-of-Experts), a disease-coordinate framework that maps multimodal evidence into a structured latent space partitioned into four distinct biological and clinical axes: genetic Risk, molecular Pathology, Neurodegeneration, and clinical Stage ($R/P/N/S$). Each observed modality parameterizes a diagonal Gaussian expert over the full RPNS vector, and a masked Product-of-Experts architecture fuses only the available modalities. Consequently, absent modalities add no factor to the fusion path, allowing the network to preserve a robust, decomposable posterior for any non-empty modality subset without synthetic imputation in the RPNS path. Through extensive missing-modality experiments on the ADNI dataset, CoPoE achieves the best all-modality performance and the highest mean AUROC across all $15$ observed-subset evaluations among standardized missing-modality fusion baselines under a shared non-PET ADNI embedding benchmark, while substantially improving raw-probability ECE, Brier score, and NLL. Furthermore, PET-supervised probing shows evidence enrichment within the pathology ($P$) block under full modalities, with tau-related signal retained even when direct fluid biospecimen inputs are withheld. Our code is available at {\normalfont\urlstyle{same}\url{https://github.com/labhai/CoPoE}}.
\end{abstract}

\begin{IEEEkeywords} Alzheimer's disease, multimodal learning, missing modality, product of experts, disease-coordinate representation \end{IEEEkeywords}

\section{Introduction}
A key challenge in multimodal Alzheimer's disease (AD) classification is that clinical categories do not fully specify amyloid/tau pathology, while the measurements that do are collected irregularly across subjects. This distinction matters because cognitively normal (CN), mild cognitive impairment (MCI), and AD labels are clinical categories, not molecular states: amyloid or tau abnormality can appear in clinically normal subjects, and the label alone does not identify the subject's biological profile \cite{jack2018niaaa,jack2024revised,jansen2015amyloid,ossenkoppele2022amyloidtau}. In this work, we use \emph{AD diagnosis} to denote computational CN/MCI/AD category classification in ADNI, not a deployable clinical diagnostic system or a new clinical diagnostic taxonomy.

In practice, this integration is complicated by irregular modality availability. Positron emission tomography (PET), cerebrospinal fluid (CSF) or blood assays, genotyping, imaging, and clinical assessments are not collected uniformly across subjects, and missingness is often systematic rather than random, reflecting cost, invasiveness, clinical indication, or site protocol \cite{nrc2012missing}. Existing missing-modality methods typically take one of two routes. They either approximate absent information before fusion through imputation or modality synthesis, or they collapse the available signals into an attention-pooled, routed-mixture, or otherwise generic latent representation. Both are limiting for biomarker-aware AD modeling: approximating absent evidence can conflate measured biomarkers with model-generated surrogates, while unstructured fusion makes it difficult to test whether the representation still preserves disease-relevant axes such as pathology and neurodegeneration.

\begin{figure}[t!]
    \centering
    \IfFileExists{data_figure.pdf}{%
        \includegraphics[width=1\linewidth]{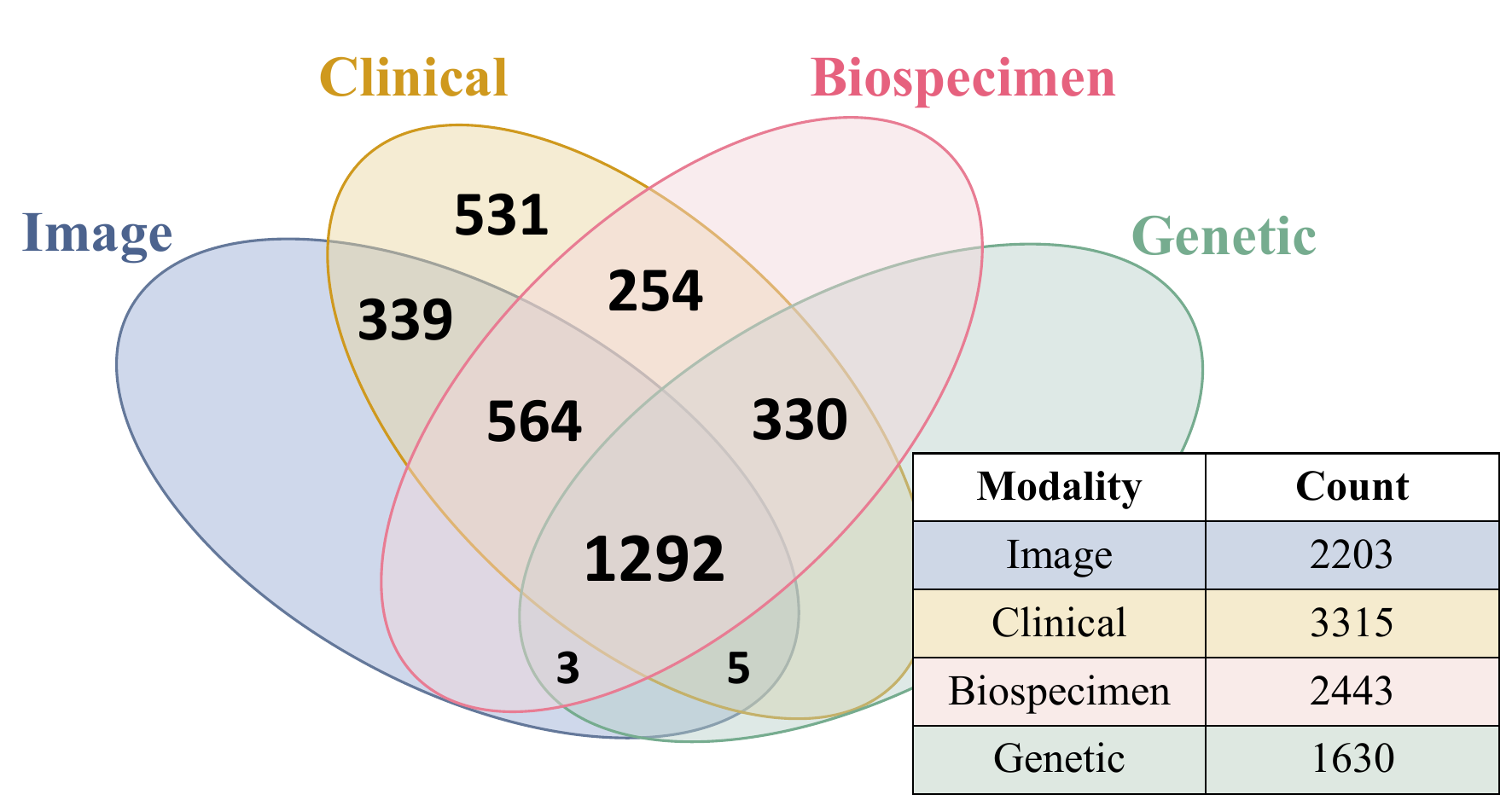}%
    }{%
        \fbox{%
        \begin{minipage}{0.92\linewidth}
        \centering
        \small
        \textbf{Placeholder: ADNI cohort and modality availability}\\[2pt]
        Cohort split and observed-subset evaluation summary.
        \end{minipage}}
    }
    \caption{ADNI modality availability. The $1{,}292$ all-four subjects form validation/test complete-case pools; training additionally uses subject-disjoint partially observed profiles.}
    \label{fig:data}
\end{figure}

We propose CoPoE, a disease-coordinate framework for missing-modality multimodal classification. CoPoE maps each observed modality to a Gaussian expert over a shared risk/pathology/neurodegeneration/stage (RPNS) latent space and multiplies only the observed experts, yielding a decomposable posterior for any non-empty modality subset without imputing missing measurements in the RPNS path. RPNS is motivated by the A/T/N biomarker scheme \cite{jack2016atn,jack2018niaaa} but used as a computational coordinate system, with each axis denoting a primary biological anchor.

The main architectural distinction is the RPNS fusion layer: rather than treating missing-modality learning as masked concatenation or feature completion, CoPoE first maps each observed modality to a Gaussian expert distribution over the same disease-coordinate space, then accumulates the observed experts through a precision-weighted product.

To make the pathology block empirically testable, PET-derived amyloid and tau heads read only the fused pathology block and are supervised with PET labels when available. ADNI is our primary benchmark cohort because it combines broad non-PET multimodal coverage (structural MRI, SNP genotypes, clinical records, biospecimen assays) with PET-derived amyloid/tau labels for probing the pathology coordinate; we evaluate CoPoE on it under all $15$ non-empty modality subsets, holding the test subjects fixed while varying the observed modality set so that subset comparisons isolate how each method uses available evidence. Even against a concurrent uncertainty-aware PoE baseline, placing observed-expert fusion inside a fixed AD-oriented coordinate space improves classification and yields a pathology coordinate that is directly testable with held-out PET labels.

CoPoE is an RPNS-structured observed-expert PoE framework rather than a generic PoE fusion model. We make three contributions:
\begin{itemize}
\item \textbf{RPNS observed-expert fusion layer.} CoPoE converts each observed modality embedding into a diagonal Gaussian expert over the full RPNS vector and fuses only observed experts by closed-form PoE. This preserves a decomposable $R/P/N/S$ posterior for any non-empty input subset without synthesizing absent modalities.

\item \textbf{PET-derived pathology probing.} Amyloid and tau positivity heads read only the fused $P$ block and are supervised by PET-derived molecular labels, allowing the pathology coordinate to be tested without using PET as an input modality.

\item \textbf{Observed-subset ADNI evaluation and component checks.} On a shared complete-case ADNI panel, we benchmark CoPoE against standardized missing-modality fusion baselines, including the concurrent PRA-PoE~\cite{yang2026prapoe}, across all $15$ non-empty observed subsets, reporting thresholded, ranking, and probability-quality metrics with component ablations that quantify each part of the RPNS-structured design.
\end{itemize}

\section{Related Work}
\label{sec:related_work}
\subsection{Missing-Modality Fusion}
Missing-modality learning spans hetero-modal aggregation, shared--specific modeling, transformer fusion, modality completion, and mixture-of-experts (MoE) routing \cite{wu2024missingsurvey,havaei2016hemis,wang2023shaspec,yao2024drfuse,yun2024flexmoe,xin2025i2moe}, which we adopt as controlled fusion baselines. Unlike CoPoE, their fused states are completion outputs, routed mixtures, interaction weights, or generic latent representations rather than disease-axis posteriors, and related AD imaging-genetics work \cite{zhou2023mclca} assumes paired modalities rather than arbitrary observed subsets.

Product-of-Experts provides a probabilistic route to observed-expert fusion: observed modalities contribute experts while absent ones do not enter the product \cite{hinton2002poe,wu2018mvae}. The closest concurrent AD-specific PoE framework, PRA-PoE, combines prototype-anchored representation alignment (PRA), feature completion, and uncertainty-aware PoE (UA-PoE) \cite{yang2026prapoe}, keeping completed features as experts. CoPoE instead adds no expert for an absent modality in the RPNS path---so a missing modality contributes no expert factor rather than a completed surrogate feature---and fuses evidence in a fixed AD-oriented coordinate space whose pathology block is testable against held-out PET labels. Because PRA-PoE uses PET as input whereas we hold PET out as labels, we adapt its official PRA and UA-PoE modules to the same frozen non-PET embeddings, splits, subsets, and metrics as all baselines, and compare against a capacity-matched flat Gaussian-PoE ablation (Sec.~\ref{sec:res_ablation}).

\subsection{Structured, Biomarker-Tested Representations}
Concept bottleneck models predict through human-defined concepts, but dense concept annotations are rare in heterogeneous AD cohorts \cite{koh2020concept}, and interpretable imaging-genetics models \cite{zhou2023gcn} generally target complete modality sets rather than arbitrary missingness. Biological frameworks such as AT(N) provide AD-relevant amyloid/tau/neurodegeneration axes but do not by themselves define a missing-modality fusion architecture \cite{jack2018niaaa,jack2024revised}, and recent work shows amyloid and tau status can be estimated from more accessible signals \cite{jasodanand2025biomarker}. CoPoE uses biomarker labels as representation tests rather than as the sole target: PET-derived amyloid and tau labels ask whether the pathology block of a classifier is biomarker-enriched, and the no-biospecimen setting separates direct fluid-biomarker evidence from weaker cross-modal pathology signal.

\section{Method}
\label{sec:method}

\begin{figure*}[t]
    \centering
    \IfFileExists{main_figure.pdf}{%
        \includegraphics[width=0.98\linewidth]{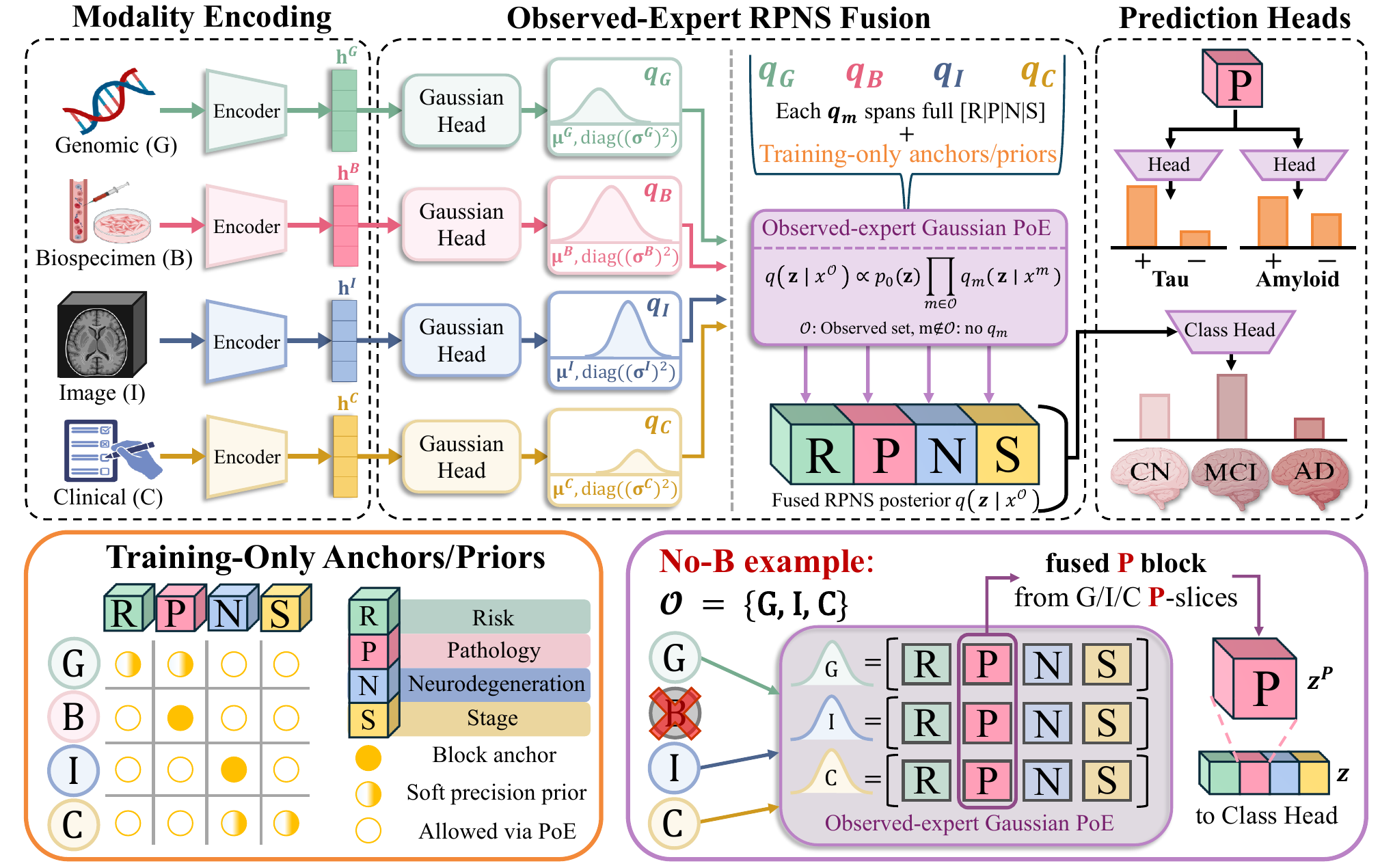}%
    }{%
        \fbox{%
        \begin{minipage}{0.92\linewidth}
        \centering\small
        \textbf{Placeholder: CoPoE architecture}\\[3pt]
        Genomic / Biospecimen / Image / Clinical inputs\\
        $\downarrow$ modality encoders $\rightarrow$ Gaussian experts over RPNS\\
        $\downarrow$ observed-expert masked PoE fusion\\
        Fused $[R|P|N|S]$ posterior $\rightarrow$ CN/MCI/AD RPNS logits\\
        Frozen modality embeddings $\rightarrow$ classification-only residual logits\\
        $P$ block $\rightarrow$ amyloid/tau pathology heads\\[2pt]
        Observed-expert RPNS fusion and PET-tested pathology readout.
        \end{minipage}}
    }
    \caption{CoPoE overview. Each observed modality emits a Gaussian expert over the full RPNS vector, and observed-expert PoE fuses only available experts. Amyloid/tau heads read the fused $P$ block; classification also uses a small residual branch over frozen embeddings. Block anchors and soft precision priors are training-only regularizers, not routing masks.}
    \label{fig:overview}
\end{figure*}

\subsection{Overview}
\label{subsec:method_overview}

CoPoE takes any non-empty subset of four modality groups: genomic ($G$), biospecimen ($B$), image ($I$), and clinical ($C$). Each available input is mapped to a Gaussian expert over a shared RPNS coordinate space, and a masked Product-of-Experts fuses these experts into a posterior over $\mathbf{z}=[\mathbf{z}^{R};\mathbf{z}^{P};\mathbf{z}^{N};\mathbf{z}^{S}]$. The fused mean drives classification, augmented by a small residual branch over the frozen embeddings, while the pathology block $\mathbf{z}^{P}$ is separately read by PET-supervised amyloid and tau heads. Fig.~\ref{fig:overview} summarizes how these RPNS coordinates, observed-expert fusion, and PET readouts are coupled.

Given frozen modality embeddings, a concatenation model would stack the available embeddings, often with zeros or masks for absent channels, and learn an unstructured classifier. CoPoE instead inserts an RPNS-structured probabilistic fusion layer: each observed embedding parameterizes a mean and variance over the same $R/P/N/S$ coordinates, and the fused posterior is obtained by multiplying only the observed Gaussian experts.

\subsection{RPNS Gaussian Experts}
\label{subsec:rpns_experts}

Each modality has an encoder $E_m$ that produces a fixed $128$-dimensional embedding $\mathbf{h}^{m}=E_m(x^m)\in\mathbb{R}^{128}$ when modality $m$ is observed. The encoders are trained before fusion and then frozen. If a modality is absent, its encoder is not evaluated and no placeholder embedding is produced for the RPNS fusion path. Input dimensions, preprocessing, and the frozen-embedding fusion interface are summarized in Sec.~\ref{sec:setup}.

CoPoE maps modality evidence into a single RPNS coordinate vector $\mathbf{z} = [\mathbf{z}^{R};\mathbf{z}^{P};\mathbf{z}^{N};\mathbf{z}^{S}] \in\mathbb{R}^{d_z}$, where the axes $R$, $P$, $N$, and $S$ anchor genetic \textbf{R}isk, molecular \textbf{P}athology, \textbf{N}eurodegeneration, and clinical \textbf{S}tage, respectively; we use these single-letter block names ($R/P/N/S$) throughout. Only the pathology coordinate is externally probed with PET labels, and the remaining blocks impose disease-oriented structure rather than independently validated variables. The main configuration uses $d_z=48$ with $R/P/N/S=8/16/16/8$, giving more capacity to pathology and neurodegeneration, the two axes most directly tied to biomarker evidence; remaining dimension choices are reported as sensitivity variants around this prespecified allocation.

For each observed modality, a Gaussian head maps $\mathbf{h}^{m}$ to a diagonal Gaussian expert over the full RPNS vector:
\begin{equation}
    q_m(\mathbf{z}\mid x^m)
    =
    \mathcal{N}\!\left(
        \boldsymbol{\mu}^m,\;
        \mathrm{diag}\big((\boldsymbol{\sigma}^m)^2\big)
    \right).
    \label{eq:gaussian_expert}
\end{equation}
The expert mean $\boldsymbol{\mu}^m$ is modality $m$'s estimate of the RPNS state, and the variance controls its coordinate-wise precision during fusion.

Each expert spans all four RPNS blocks rather than only its most direct block, which is central under missingness: when biospecimen is absent, the pathology block is inferred from the $P$-coordinates of the remaining experts rather than left empty. Biospecimen and imaging provide the most direct $P$ and $N$ evidence, while genomic and clinical features carry indirect evidence through risk, progression, and severity; Sec.~\ref{sec:res_bio} probes how much $P$-signal persists when biospecimen is removed.

\subsection{Observed-Expert Masked Gaussian PoE Fusion}
\label{subsec:masked_poe}

Let $\mathcal{O}\subseteq\{G,B,I,C\}$ be the observed modality set for a subject. CoPoE fuses only the corresponding experts:
\begin{equation}
    q(\mathbf{z}\mid x^{\mathcal{O}})
    \propto
    p_0(\mathbf{z})
    \prod_{m\in\mathcal{O}} q_m(\mathbf{z}\mid x^m),
    \label{eq:masked_poe}
\end{equation}
where $p_0(\mathbf{z})=\mathcal{N}(\mathbf{0}, \mathrm{diag}(\boldsymbol{\sigma}_0^2))$ is a zero-mean diagonal Gaussian prior with learnable per-coordinate variance. Missing modalities contribute no factor to Eq.~\eqref{eq:masked_poe}; they are not imputed, synthesized, or represented by learned missing tokens in the fused RPNS posterior.

Because the prior and experts are diagonal Gaussians, fusion is closed-form in each coordinate. Omitting the coordinate index and writing $\lambda^m=1/(\sigma^m)^2$ and $\lambda_0=1/\sigma_0^2$, the fused precision, mean, and variance are
\begin{equation}
\begin{aligned}
    \lambda^{\star}
    &=
    \lambda_0+\sum_{m\in\mathcal{O}}\lambda^m, \\
    \mu^{\star}
    &=
    \frac{\sum_{m\in\mathcal{O}}\lambda^m\mu^m}{\lambda^{\star}},
    \qquad
    (\sigma^{\star})^2
    =
    \frac{1}{\lambda^{\star}} .
\end{aligned}
    \label{eq:poe_moment}
\end{equation}
Thus, the fused mean is a prior-regularized precision-weighted combination of the observed experts: the zero-mean prior shrinks the estimate when few experts are present, and adding an observed expert increases fused precision where that expert is confident. Since the product is computed in the RPNS coordinate system, the fused posterior remains decomposable into the same $R/P/N/S$ blocks for every non-empty subset.

\subsection{Block Alignment and Pathology Readout}
\label{subsec:anchoring_validation}

To encourage biologically plausible modality-block specialization, we add a soft block-alignment loss for two pairs $\mathcal{A}=\{(I,N),(B,P)\}$, linking image evidence to neurodegeneration and biospecimen evidence to molecular pathology; genomic and clinical modalities are not hard-assigned, being broader signals that affect several axes.

For each specialist pair $(m,b)\in\mathcal{A}$, we align the modality-specific block mean $\boldsymbol{\mu}^{m,b}$ with the fused block mean $\boldsymbol{\mu}^{\star,b}$ using an in-batch contrastive objective. Let $\mathbf{a}^{m,b}_n$ and $\mathbf{f}^{b}_n$ denote L2-normalized projections of these two block means, and define $s^{m,b}_{n,n'}=\langle\mathbf{a}^{m,b}_n,\mathbf{f}^{b}_{n'}\rangle$. The loss is
\begin{equation}
    \mathcal{L}_{\mathrm{align}}
    =
    \sum_{(m,b)\in\mathcal{A}}
    -\frac{1}{|\mathcal{B}_m|}
    \sum_{n\in\mathcal{B}_m}
    \log
    \frac{\exp(s^{m,b}_{n,n}/\tau)}
    {\sum_{n'\in\mathcal{B}_m}\exp(s^{m,b}_{n,n'}/\tau)},
    \label{eq:alignment}
\end{equation}
where $\mathcal{B}_m$ is the set of in-batch subjects with modality $m$ observed and $\tau$ is a temperature; pairs with $|\mathcal{B}_m|=0$ in a given batch are skipped. Block specialization is thus encouraged, while the independent test of molecular pathology comes from PET-derived amyloid and tau labels.

Two binary pathology heads read only the fused $P$ block:
\begin{equation}
\begin{aligned}
    \hat{y}^{\mathrm{amy}}
    &=
    \mathrm{sigmoid}\big(g_{\mathrm{amy}}(\boldsymbol{\mu}^{\star,P})\big), \\
    \hat{y}^{\mathrm{tau}}
    &=
    \mathrm{sigmoid}\big(g_{\mathrm{tau}}(\boldsymbol{\mu}^{\star,P})\big) ,
\end{aligned}
    \label{eq:pathology_heads}
\end{equation}
where $\mathrm{sigmoid}(\cdot)$ is the logistic function and each $g$ is a lightweight head on the fused pathology block. They are trained only for subjects with the corresponding PET label and evaluated as molecular readouts from $P$, not as alternative CN/MCI/AD classifications. Because biospecimen inputs can contain fluid amyloid/tau assays whereas the labels are PET-derived, Sec.~\ref{sec:res_bio} reports both all-modality and no-biospecimen results.

\subsection{Classification, Training, and Inference}
\label{subsec:classification_training_inference}

The fused RPNS mean $\boldsymbol{\mu}^{\star}$ is layer-normalized (LN) and passed to a compact classifier. To avoid forcing all classification information through the compact RPNS bottleneck, CoPoE also includes a small classification-only residual branch over the frozen modality embeddings:
\begin{equation}
    \boldsymbol{\ell}
    =
    f_{\mathrm{RPNS}}\big(\mathrm{LN}(\boldsymbol{\mu}^{\star})\big)
    +
    \alpha\,
    f_{\mathrm{res}}
    \big([\tilde{\mathbf{h}}^{G};
          \tilde{\mathbf{h}}^{B};
          \tilde{\mathbf{h}}^{I};
          \tilde{\mathbf{h}}^{C}]\big),
    \label{eq:classification_logits}
\end{equation}
where $\boldsymbol{\ell}$ are the CN/MCI/AD logits, $\tilde{\mathbf{h}}^{m}=\mathbf{h}^{m}$ if $m\in\mathcal{O}$ and $\tilde{\mathbf{h}}^{m}=\mathbf{0}$ otherwise, and $\alpha=0.5$ in all experiments. This zero vector is a fixed-width zero-fill convention for the residual branch only; it does not enter the RPNS product in Eq.~\eqref{eq:masked_poe} and is never read by the amyloid or tau heads.

The fusion module is trained by minimizing
\begin{equation}
\begin{aligned}
\mathcal{L}={}&
\mathcal{L}_{\mathrm{cls}}
+\lambda_{\mathrm{amy}}\mathcal{L}_{\mathrm{amy}}
+\lambda_{\mathrm{tau}}\mathcal{L}_{\mathrm{tau}} \\
&+\lambda_{\mathrm{KL}}\mathcal{L}_{\mathrm{KL}}
+\lambda_{\mathrm{align}}\mathcal{L}_{\mathrm{align}} \\
&+\lambda_{\mathrm{sub}}\mathcal{L}_{\mathrm{sub}}
+\lambda_{\mathrm{casc}}\mathcal{L}_{\mathrm{casc}} .
\end{aligned}
\label{eq:total_loss}
\end{equation}
Here, $\mathcal{L}_{\mathrm{cls}}$ is the classification cross-entropy. $\mathcal{L}_{\mathrm{amy}}$ and $\mathcal{L}_{\mathrm{tau}}$ are positive-class-weighted focal binary cross-entropy losses evaluated only for subjects with the corresponding PET label. $\mathcal{L}_{\mathrm{KL}}$ is a Kullback--Leibler (KL) term that regularizes the per-modality and fused Gaussian posteriors toward the prior. $\mathcal{L}_{\mathrm{align}}$ is the block-alignment loss in Eq.~\eqref{eq:alignment}. $\mathcal{L}_{\mathrm{sub}}$ applies the classification cross-entropy under a random non-empty subset of the available modalities, against the original label.

The final term, $\mathcal{L}_{\mathrm{casc}}$, is a low-weight cascade prior that converts each generalist expert's block-wise precision mass into a distribution over $[R,P,N,S]$ and biases it toward a fixed target via $D_{\mathrm{KL}}(P_m\Vert\pi_m)$: $\pi_G=[0.40,0.40,0.10,0.10]$ and $\pi_C=[0.05,0.20,0.30,0.45]$; the specialists $I$ and $B$ are instead handled by the anchors $I{\rightarrow}N$ and $B{\rightarrow}P$. We use this as a fixed implementation regularizer and do not treat it as a separate methodological contribution.

At inference, only the encoders of observed modalities are evaluated and only their experts enter Eq.~\eqref{eq:masked_poe}, so the same fusion rule applies to any subset with no network change or separate imputation step; the model returns class probabilities, the fused RPNS posterior, and optional amyloid/tau probabilities from the $P$ block.

\section{Experiments and Results}
\label{sec:experiments}

\begin{table*}[!t]
\caption{All-modality CN/MCI/AD results. BAcc/macro-F1/AUROC are \%; ECE/Brier/NLL raw. Mean $\pm$ std over five seeds; bold/underline mark best/second; stars mark Holm-corrected paired patient-level bootstrap significance of CoPoE over that baseline ($^\ast$$p{<}0.05$, $^{\ast\ast}$$p{<}0.01$).}
\label{tab:allfour_summary}
\centering
\footnotesize
\setlength{\tabcolsep}{4.1pt}
\renewcommand{\arraystretch}{1.05}
\resizebox{0.90\textwidth}{!}{%
\begin{tabular}{lcccccc}
\toprule
\textbf{Model} & \textbf{BAcc$\uparrow$} & \textbf{macro-F1$\uparrow$} & \textbf{AUROC$\uparrow$} & \textbf{ECE$\downarrow$} & \textbf{Brier$\downarrow$} & \textbf{NLL$\downarrow$} \\
\midrule
mmFormer~{\tiny\cite{zhang2022mmformer}} & $63.26{\pm}1.05^{\ast\ast}$ & $63.20{\pm}1.13^{\ast\ast}$ & $81.50{\pm}0.92^{\ast\phantom{\ast}}$ & $0.220{\pm}0.044^{\ast\phantom{\ast}}$ & $0.545{\pm}0.039^{\ast\ast}$ & $1.153{\pm}0.355^{\ast\ast}$ \\
ShaSpec~{\tiny\cite{wang2023shaspec}} & $64.40{\pm}1.26^{\ast\ast}$ & $64.37{\pm}1.46^{\ast\ast}$ & $82.39{\pm}0.53^{\ast\ast}$ & $0.178{\pm}0.033^{\phantom{\ast\ast}}$ & $0.497{\pm}0.030^{\ast\ast}$ & $0.909{\pm}0.100^{\ast\ast}$ \\
DrFuse~{\tiny\cite{yao2024drfuse}} & $61.15{\pm}0.61^{\ast\ast}$ & $61.30{\pm}1.21^{\ast\ast}$ & $79.69{\pm}0.71^{\ast\ast}$ & $0.242{\pm}0.030^{\ast\ast}$ & $0.582{\pm}0.037^{\ast\ast}$ & $1.444{\pm}0.425^{\ast\ast}$ \\
MUSE~{\tiny\cite{wu2024muse}} & $60.65{\pm}1.37^{\ast\ast}$ & $61.44{\pm}0.92^{\ast\ast}$ & $76.68{\pm}1.17^{\ast\ast}$ & $0.175{\pm}0.048^{\phantom{\ast\ast}}$ & $0.537{\pm}0.017^{\ast\ast}$ & $1.028{\pm}0.123^{\ast\ast}$ \\
Flex-MoE~{\tiny\cite{yun2024flexmoe}} & $64.49{\pm}0.62^{\ast\phantom{\ast}}$ & $62.99{\pm}0.46^{\ast\ast}$ & $82.14{\pm}0.66^{\phantom{\ast\ast}}$ & $\underline{0.149{\pm}0.021}^{\phantom{\ast\ast}}$ & $0.491{\pm}0.015^{\ast\phantom{\ast}}$ & $\underline{0.861{\pm}0.062}^{\ast\ast}$ \\
$I^2$MoE~{\tiny\cite{xin2025i2moe}} & $62.90{\pm}1.13^{\ast\phantom{\ast}}$ & $61.53{\pm}1.23^{\ast\ast}$ & $82.16{\pm}1.27^{\phantom{\ast\ast}}$ & $0.235{\pm}0.038^{\ast\ast}$ & $0.547{\pm}0.030^{\ast\ast}$ & $1.331{\pm}0.255^{\ast\ast}$ \\
MoE-Ret~{\tiny\cite{moeretriever2025}} & $62.63{\pm}1.10^{\ast\ast}$ & $61.94{\pm}2.65^{\ast\ast}$ & $81.33{\pm}2.86^{\phantom{\ast\ast}}$ & $0.229{\pm}0.057^{\phantom{\ast\ast}}$ & $0.555{\pm}0.076^{\ast\phantom{\ast}}$ & $1.321{\pm}0.450^{\ast\ast}$ \\
PRA-PoE~{\tiny\cite{yang2026prapoe}} & $\underline{66.45{\pm}1.24}^{\ast\phantom{\ast}}$ & $\underline{65.75{\pm}0.96}^{\ast\ast}$ & $\underline{83.72{\pm}0.64}^{\ast\phantom{\ast}}$ & $0.157{\pm}0.088^{\ast\phantom{\ast}}$ & $\underline{0.482{\pm}0.071}^{\ast\ast}$ & $0.957{\pm}0.043^{\ast\ast}$ \\
\rowcolor{oursorange}
\textbf{CoPoE}~{\tiny(Ours)} & $\mathbf{71.87{\pm}0.89}^{\phantom{\ast\ast}}$ & $\mathbf{71.92{\pm}0.62}^{\phantom{\ast\ast}}$ & $\mathbf{85.80{\pm}0.33}^{\phantom{\ast\ast}}$ & $\mathbf{0.086{\pm}0.011}^{\phantom{\ast\ast}}$ & $\mathbf{0.408{\pm}0.003}^{\phantom{\ast\ast}}$ & $\mathbf{0.681{\pm}0.006}^{\phantom{\ast\ast}}$ \\
\bottomrule
\end{tabular}%
}
\end{table*}

\subsection{Experimental Setup}
\label{sec:setup}

\textbf{Evaluation frame.} All reported claims are evaluated within a standardized ADNI missing-modality fusion benchmark. \emph{AD diagnosis} denotes computational CN/MCI/AD classification, not a deployable clinical system or a new taxonomy; \emph{state-of-the-art} refers to comparisons against standardized fusion baselines that share the same frozen non-PET embeddings, subject-disjoint splits, and observed-subset evaluations. \emph{Robustness} refers to stability across the $15$ observed-modality subsets of a common complete-case panel, with training additionally exposed to naturally incomplete profiles. \emph{Pathology capture} is assessed by PET-supervised $P$-block readouts against matched readout controls, distinguishing retained signal from exclusive localization.

We evaluate CoPoE on the Alzheimer's Disease Neuroimaging Initiative (ADNI) \cite{jack2008adni} for three-class classification using four modality groups: genomic ($G$), biospecimen ($B$), image ($I$), and clinical ($C$); Fig.~\ref{fig:data} summarizes modality availability across the cohort. Of the $1{,}292$ all-four subjects, $100$/$200$ are held out for validation/test; training uses $3{,}018$ subject-disjoint profiles ($992$ non-held-out all-four plus $2{,}026$ naturally partially observed). The CN/MCI/AD split is $1{,}370/1{,}136/512$ ($45.4/37.6/17.0\%$) over training, with matched minority-AD proportions in validation and test ($15.0\%$, $14.5\%$); we therefore report balanced accuracy and macro-F1, with each subject in exactly one participant-level split and modality encoders pretrained on the training split only (no held-out subject or visit in encoder pretraining). For the all-subset evaluation, the complete-case validation/test subjects serve as a common evaluation panel: each row of Table~\ref{tab:allsubset_auroc} provides only the listed observed modalities to every method, while the auxiliary subset loss during training samples random non-empty subsets only from modalities naturally available for each subject.

\textbf{Input modalities and fusion interface.} We use four non-PET ADNI input modalities: T1-weighted structural MRI (sMRI)-derived image volumes ($I$), SNP genotype dosages ($G$), clinical tabular records ($C$), and biospecimen/CSF assays ($B$). MRI was intensity-corrected, MUSE-segmented, registered to MNI space, and resampled/cropped to $96{\times}112{\times}96$. Genotypes from ADNI\,1, ADNI\,GO/2, and ADNI3 were lifted to NCBI build 38/UCSC hg38, merged across batches, LD-pruned with $(50,5,0.1)$, encoded as $0/1/2$ dosages, and reduced to the top $16{,}384$ SNPs ranked by a chi-square association test fit on the training split only. Clinical variables are drawn from medical history, neurological examination, demographic, medication, and vital-sign tables---not from cognitive or functional composites such as MMSE, CDR, ADAS-Cog, or FAQ---and exclude diagnosis-defining fields such as PTCOGBEG, PTADDX, and PTADBEG, so the clinical modality does not reconstruct the CN/MCI/AD label from its defining instruments. Biospecimen variables come from Roche Elecsys CSF/biofluid assays, including ABETA40, ABETA42, total tau, and phosphorylated tau, together with APOE genotype one-hot fields (GENOTYPE\_*), so APOE-related risk is represented explicitly in $B$, while $G$ contributes genome-wide SNP dosages that include regional variation near the APOE/TOMM40 locus rather than the defining $\varepsilon$ variants themselves; we keep both native encodings rather than deduplicating, and no-$B$ drops the explicit biospecimen APOE fields with the CSF assays while $G$ retains its genome-wide SNP evidence. Numerical variables were min--max scaled to $[-1,1]$, categorical variables one-hot encoded, and SNP feature selection and imputation statistics were fit on the training split only, using no test-set information. For a standardized fusion benchmark, all methods receive the same frozen $128$-d modality embeddings from lightweight modality encoders; these encoders define the common input interface rather than the claimed contribution.

\textbf{PET-derived molecular labels.} Amyloid and tau are not input modalities; they supervise only the $P$-block readout heads and do not alter the class labels. Amyloid positivity is defined from AV45/Florbetapir PET cortical amyloid burden using ADNI/UC~Berkeley mean cortical SUVR with cerebellar reference (MCSUVRCERE $>1.11$, approximately $20$ Centiloids) \cite{royse2021centiloid}. Tau positivity is defined from FTP/Flortaucipir (AV-1451) PET meta-temporal burden using META\_TEMPORAL\_SUVR $>1.27$ \cite{cho2023leads}. Thus, biospecimen inputs contain CSF/biofluid assays, whereas amyloid/tau targets are PET-defined molecular labels; no PET-derived feature is provided to CoPoE or any baseline at fusion time. The test set contains $114$ amyloid-labeled subjects ($58/56$ positive/negative) and $91$ tau-labeled subjects ($30/61$). The broad training set contains $677$ amyloid- and $820$ tau-labeled subjects ($560$ and $409$ among the strict all-four training subjects). We report a no-biospecimen (no-$B$) condition to separate PET readout behavior from direct fluid-biomarker input.

\begin{table*}[!t]
\caption{Observed-subset AUROC on the common complete-case panel. $\bullet$\,=\,observed, $\times$\,=\,withheld; Average is unweighted over the $15$ subsets; bold/underline mark best/second.}
\label{tab:allsubset_auroc}
\centering
\vspace{-0.35em}
{\fontsize{6.05}{6.75}\selectfont
\setlength{\tabcolsep}{3.0pt}
\renewcommand{\arraystretch}{1.0}
\setlength{\aboverulesep}{0.2ex}
\setlength{\belowrulesep}{0.2ex}
\resizebox{\textwidth}{!}{%
\begin{tabular}{cc|MMMM|cccccccc>{\columncolor{oursorange}}c}
\toprule
\multicolumn{2}{c|}{\multirow{2}{*}{\textbf{Settings}}} &
\multicolumn{4}{c|}{\parbox[c]{6.5em}{\centering\textbf{Observed Modalities}\\[-1pt]\textbf{at Test Time}}} &
\multicolumn{9}{c}{\textbf{AUROC (\%)}} \\
\cmidrule(lr){3-6}\cmidrule(lr){7-15}
\multicolumn{2}{c|}{} & \textbf{G} & \textbf{B} & \textbf{I} & \textbf{C}
& \modelhdr{mmFormer}{\cite{zhang2022mmformer}}
& \modelhdr{ShaSpec}{\cite{wang2023shaspec}}
& \modelhdr{DrFuse}{\cite{yao2024drfuse}}
& \modelhdr{MUSE}{\cite{wu2024muse}}
& \modelhdr{Flex-MoE}{\cite{yun2024flexmoe}}
& \modelhdr{$I^2$MoE}{\cite{xin2025i2moe}}
& \modelhdr{MoE-Ret}{\cite{moeretriever2025}}
& \modelhdr{PRA-PoE}{\cite{yang2026prapoe}}
& \begin{tabular}[c]{@{}c@{}}\textbf{CoPoE}\\[-1.5pt]{\tiny (Ours)}\end{tabular} \\
\midrule
\multirow{14}{*}{\rotatebox[origin=c]{90}{\textbf{Withheld}}}
 & 1 & \obs & \miss & \miss & \miss & \auc{63.34}{1.89} & \auc{63.30}{3.44} & \auc{61.75}{3.07} & \auc{62.96}{2.73} & \auc{59.71}{1.54} & \auc{62.93}{1.21} & \auc{55.03}{2.10} & \aucsecond{65.99}{1.61} & \aucbest{67.64}{1.36} \\
 & 2 & \miss & \obs & \miss & \miss & \auc{70.96}{1.50} & \auc{68.11}{2.16} & \auc{69.39}{1.78} & \aucbest{72.68}{0.64} & \auc{67.22}{2.29} & \auc{59.40}{2.53} & \auc{69.97}{1.14} & \auc{70.74}{1.27} & \aucsecond{71.99}{1.18} \\
 & 3 & \miss & \miss & \obs & \miss & \auc{70.24}{2.31} & \aucbest{72.33}{1.29} & \auc{71.13}{2.02} & \auc{70.84}{1.48} & \auc{67.29}{0.66} & \auc{69.24}{2.79} & \auc{68.88}{3.32} & \aucsecond{71.83}{0.51} & \auc{69.91}{1.47} \\
 & 4 & \miss & \miss & \miss & \obs & \auc{67.31}{2.72} & \aucsecond{70.69}{1.03} & \auc{65.61}{3.30} & \auc{69.38}{1.86} & \auc{67.78}{0.78} & \auc{64.49}{2.56} & \auc{65.80}{1.58} & \auc{66.45}{1.84} & \aucbest{70.79}{1.26} \\
\cmidrule(l){2-15}
 & 5 & \obs & \obs & \miss & \miss & \auc{73.63}{0.98} & \auc{74.55}{1.03} & \auc{72.26}{1.13} & \auc{70.47}{0.81} & \auc{70.34}{1.28} & \auc{67.53}{2.31} & \auc{71.77}{1.97} & \aucsecond{75.82}{2.12} & \aucbest{76.61}{0.27} \\
 & 6 & \obs & \miss & \obs & \miss & \auc{75.19}{1.81} & \aucsecond{76.89}{1.33} & \auc{72.76}{1.44} & \auc{69.74}{2.26} & \auc{64.19}{1.96} & \auc{71.61}{3.29} & \auc{69.63}{2.66} & \auc{76.42}{0.70} & \aucbest{79.56}{0.29} \\
 & 7 & \obs & \miss & \miss & \obs & \auc{70.58}{2.43} & \auc{73.23}{0.97} & \auc{68.59}{1.49} & \auc{67.99}{1.91} & \auc{69.80}{2.35} & \auc{68.67}{2.27} & \auc{67.28}{2.49} & \aucsecond{74.26}{1.89} & \aucbest{75.48}{0.64} \\
 & 8 & \miss & \obs & \obs & \miss & \auc{80.16}{1.11} & \aucsecond{81.30}{1.02} & \auc{80.81}{0.69} & \auc{79.41}{1.36} & \auc{72.47}{1.60} & \auc{71.49}{1.95} & \auc{75.82}{2.27} & \auc{79.30}{1.29} & \aucbest{82.36}{0.25} \\
 & 9 & \miss & \obs & \miss & \obs & \auc{74.19}{1.08} & \aucsecond{76.73}{1.37} & \auc{74.49}{1.39} & \auc{75.79}{0.89} & \auc{73.81}{0.98} & \auc{66.26}{3.22} & \auc{76.51}{1.09} & \auc{73.46}{1.07} & \aucbest{76.99}{0.83} \\
 & 10 & \miss & \miss & \obs & \obs & \auc{77.20}{2.41} & \aucbest{79.32}{0.53} & \auc{75.35}{2.46} & \auc{77.01}{1.48} & \auc{75.12}{2.02} & \auc{73.92}{1.83} & \auc{75.69}{1.65} & \auc{75.53}{1.48} & \aucsecond{79.00}{0.76} \\
\cmidrule(l){2-15}
 & 11 & \obs & \obs & \obs & \miss & \auc{80.02}{0.72} & \auc{81.14}{0.67} & \auc{78.35}{0.94} & \auc{74.97}{0.87} & \auc{73.09}{1.46} & \auc{77.75}{1.88} & \auc{77.21}{2.69} & \aucsecond{82.43}{0.96} & \aucbest{84.65}{0.37} \\
 & 12 & \obs & \obs & \miss & \obs & \auc{76.58}{1.09} & \auc{77.97}{0.53} & \auc{75.28}{1.05} & \auc{73.35}{1.17} & \auc{76.59}{0.58} & \auc{74.21}{1.81} & \auc{76.33}{1.26} & \aucsecond{78.48}{0.89} & \aucbest{79.46}{0.21} \\
 & 13 & \obs & \miss & \obs & \obs & \auc{78.48}{1.99} & \auc{79.59}{0.76} & \auc{75.59}{0.84} & \auc{72.75}{1.79} & \auc{76.93}{1.52} & \auc{76.66}{0.84} & \auc{76.13}{1.85} & \aucsecond{80.86}{1.65} & \aucbest{82.82}{0.36} \\
 & 14 & \miss & \obs & \obs & \obs & \auc{81.59}{0.98} & \aucsecond{83.01}{1.19} & \auc{81.10}{1.45} & \auc{80.78}{0.83} & \auc{78.52}{2.43} & \auc{75.36}{1.73} & \auc{81.66}{2.22} & \auc{80.17}{1.33} & \aucbest{84.21}{0.29} \\
\midrule
\textbf{Full} & 15 & \obs & \obs & \obs & \obs & \auc{81.50}{0.92} & \auc{82.39}{0.53} & \auc{79.69}{0.71} & \auc{76.68}{1.17} & \auc{82.14}{0.66} & \auc{82.16}{1.27} & \auc{81.33}{2.86} & \aucsecond{83.72}{0.64} & \aucbest{85.80}{0.33} \\
\midrule
\multicolumn{6}{c|}{\textbf{Average}} & 74.73 & \underline{76.04} & 73.48 & 72.99 & 71.67 & 70.78 & 72.60 & 75.70 & \textbf{77.82} \\
\bottomrule
\end{tabular}%
}
}
\vspace{-0.4em}
\end{table*}

\textbf{Evaluation.} Classification is measured by BAcc, macro-F1, and macro one-vs-rest AUROC; output-level probability quality by expected calibration error (ECE, $15$-bin)~\cite{guo2017calibration}, Brier score, and NLL; and PET readouts by AUROC/AUPRC. Results are reported as mean $\pm$ std over five seeds. Significance in Table~\ref{tab:allfour_summary} and for the PET readouts is assessed by a patient-level paired bootstrap over the fixed test cohort with Holm correction within each metric, reflecting cohort sampling rather than training-initialization variability. For the $15$-subset AUROC comparison, subset means are treated as paired blocks and tested with one-sided Wilcoxon signed-rank tests with Holm correction.

\textbf{Baselines.} We compare against eight standardized fusion-rule adaptations: mmFormer, ShaSpec, DrFuse, MUSE, Flex-MoE, $I^2$MoE, MoE-Ret, and PRA-PoE (the closest concurrent AD-specific PoE method). All methods consume the same frozen embeddings, splits, observed subsets, five seeds, training budget, and random-subset training augmentation while retaining their published fusion and missing-modality handling, isolating fusion behavior from encoder, preprocessing, and training-protocol differences. For PRA-PoE, the PRA and UA-PoE auxiliary losses and $10$-sample Monte Carlo prediction are retained, but PET is held out as a label rather than an input, so no method receives PET-derived features at fusion time. Model selection uses mean validation BAcc over the $15$ observed subsets, and all-subset AUROC is the primary threshold-independent missing-modality comparison.

\textbf{Implementation.} CoPoE uses $d_z=48$ with RPNS allocation $8/16/16/8$ and embedding width $128$. The fusion module is trained for up to $50$ epochs with batch size $64$, AdamW at learning rate $7.5{\times}10^{-5}$, and a cosine schedule, selecting on validation BAcc. The residual branch uses scale $\alpha=0.5$. Loss weights are $\lambda_{\mathrm{KL}}=5{\times}10^{-4}$, $\lambda_{\mathrm{align}}=1.0$ (temperature $\tau=0.2$, specialist pairs $I{:}N$ and $B{:}P$), $\lambda_{\mathrm{amy}}=0.3$, $\lambda_{\mathrm{tau}}=0.7$, $\lambda_{\mathrm{sub}}=0.25$, and $\lambda_{\mathrm{casc}}=0.1$ over the two generalist experts $G$ and $C$. Classification uses inverse-frequency CE. Amyloid/tau BCE uses training-fold $w_+=n_{\mathrm{neg}}/n_{\mathrm{pos}}$ and a focal factor $(1-p_t)^\gamma$ with fixed $\gamma=1$, where $p_t=y\hat p+(1-y)(1-\hat p)$. Cascade targets are fixed; no validation/test labels set these quantities. Probability-quality metrics use raw predicted probabilities without post-hoc calibration.

\subsection{All-Modality Multi-Metric Overview}
\label{sec:res_allfour}

Table~\ref{tab:allfour_summary} gives the all-modality overview across thresholded classification, ranking, and output-level probability quality. CoPoE is best on all six metrics. Against the strongest non-CoPoE baseline for each metric, CoPoE improves BAcc by $5.42$ percentage points, macro-F1 by $6.17$ percentage points, and AUROC by $2.08$ percentage points, while reducing ECE, Brier score, and NLL by $0.063$, $0.074$, and $0.180$, respectively. Patient-level bootstrap tests with Holm correction (Table~\ref{tab:allfour_summary}) show significant BAcc, macro-F1, Brier, and NLL gains over all baselines and AUROC gains over most. Since PRA-PoE already uses uncertainty-aware Gaussian PoE, these gains support the RPNS-structured fusion design beyond generic Gaussian PoE; Sec.~\ref{sec:res_ablation} separates this from the residual branch and class-balanced training. We interpret ECE, Brier score, and NLL as output-level calibration and probability-quality metrics of the final predictive distribution, not as evidence that the latent Gaussian expert variances are themselves calibrated uncertainty estimates.

\subsection{Observed-Subset AUROC}
\label{sec:res_main}

Table~\ref{tab:allsubset_auroc} evaluates all $15$ non-empty observed-modality subsets on a common complete-case test panel, where each row uses the same subjects so differences reflect available input evidence rather than cohort composition. CoPoE achieves the highest AUROC in $12$ of the $15$ subset settings and the best all-subset average ($77.82\%$ vs.\ $76.04\%$ for the strongest baseline). The three exceptions occur in single- or two-modality settings (biospecimen-only for MUSE; image-only and image+clinical for ShaSpec), indicating that CoPoE is not uniformly dominant with minimal evidence but gains as complementary evidence accumulates. Since models are selected on validation BAcc, this AUROC ranking is not the optimized quantity. Treating subset means as paired blocks, CoPoE exceeds every baseline after Holm-corrected one-sided Wilcoxon signed-rank tests; against the strongest baseline, ShaSpec, the mean AUROC gain is $1.78$ percentage points ($p_{\mathrm{Holm}}=2.1{\times}10^{-3}$).

\subsection{PET Biomarker Readout Analysis}
\label{sec:res_bio}

\begin{table*}[t]
\caption{PET readouts by RPNS coordinate. Labels are PET-derived; no-$B$ withholds biospecimen. AUROC/AUPRC \%, mean $\pm$ std over five seeds; bold marks best.}
\label{tab:bio_probe}
\centering
\scriptsize
\setlength{\tabcolsep}{3.0pt}
\resizebox{\textwidth}{!}{%
\begin{tabular}{l|cc|cc|cc|cc}
\toprule
& \multicolumn{2}{c|}{\textbf{All modalities: Amyloid}} &
  \multicolumn{2}{c|}{\textbf{All modalities: Tau}} &
  \multicolumn{2}{c|}{\textbf{No-$B$: Amyloid}} &
  \multicolumn{2}{c}{\textbf{No-$B$: Tau}} \\
\cmidrule(lr){2-3}\cmidrule(lr){4-5}\cmidrule(lr){6-7}\cmidrule(lr){8-9}
\textbf{Readout source}
& \textbf{AUROC} & \textbf{AUPRC}
& \textbf{AUROC} & \textbf{AUPRC}
& \textbf{AUROC} & \textbf{AUPRC}
& \textbf{AUROC} & \textbf{AUPRC} \\
\midrule
$R$ block
& $75.26{\pm}5.58$ & $76.98{\pm}7.25$
& $68.45{\pm}4.63$ & $53.15{\pm}6.19$
& $66.55{\pm}3.02$ & $66.28{\pm}5.17$
& $64.77{\pm}5.64$ & $51.07{\pm}6.34$ \\
$P$ block
& $\mathbf{91.83{\pm}0.70}$ & $\mathbf{92.82{\pm}0.89}$
& $\mathbf{75.25{\pm}1.18}$ & $\mathbf{64.66{\pm}2.00}$
& $68.90{\pm}2.68$ & $69.83{\pm}4.99$
& $\mathbf{67.57{\pm}2.32}$ & $\mathbf{56.27{\pm}1.61}$ \\
$N$ block
& $70.27{\pm}10.00$ & $70.00{\pm}11.52$
& $66.44{\pm}4.35$ & $50.57{\pm}5.77$
& $64.00{\pm}1.59$ & $63.27{\pm}2.27$
& $56.96{\pm}7.28$ & $45.23{\pm}4.98$ \\
$S$ block
& $83.87{\pm}3.89$ & $86.74{\pm}3.34$
& $70.02{\pm}8.66$ & $58.21{\pm}8.22$
& $68.53{\pm}3.80$ & $70.17{\pm}3.35$
& $63.01{\pm}6.60$ & $47.90{\pm}7.15$ \\
\midrule
Full RPNS
& $89.58{\pm}2.57$ & $91.56{\pm}2.14$
& $72.55{\pm}3.50$ & $62.50{\pm}3.80$
& $67.67{\pm}0.93$ & $66.42{\pm}1.78$
& $61.37{\pm}8.31$ & $49.57{\pm}5.93$ \\
CN/MCI/AD logits
& $85.33{\pm}0.48$ & $88.42{\pm}0.45$
& $72.48{\pm}2.21$ & $58.67{\pm}4.38$
& $\mathbf{71.32{\pm}0.83}$ & $\mathbf{72.37{\pm}0.43}$
& $65.77{\pm}1.09$ & $52.61{\pm}3.08$ \\
\bottomrule
\end{tabular}%
}
\end{table*}

We test whether the pathology coordinate carries molecular pathology signal; since the block is supervised by PET-derived labels (not by the CSF biospecimen inputs), we compare its readout against those from $R$, $N$, $S$, the full RPNS vector, and the CN/MCI/AD logits under the same held-out PET labels and probe, trained only on the training split. AUPRC should be read against the positive prevalence in the labeled test subset ($50.9\%$ amyloid, $33.0\%$ tau). These probes control for readout architecture and label availability, not for where PET supervision is placed, so the test asks whether PET-related signal is enriched in the pathology block rather than whether that block emerges unsupervised.

The all-modality results in Table~\ref{tab:bio_probe} support this intended enrichment. The pathology block is the strongest readout for both amyloid ($91.83$ AUROC, $92.82$ AUPRC) and tau ($75.25$ AUROC, $64.66$ AUPRC), outperforming both the full RPNS vector and the CN/MCI/AD logits. The no-$B$ setting separates signal \emph{retention} from exclusive localization: tau remains strongest in the pathology block, retaining tau-related signal after the most direct fluid-biomarker evidence is removed, whereas amyloid is then best predicted from the CN/MCI/AD logits. The no-$B$ $P$-block amyloid readout ($68.90$ AUROC) nonetheless remains above chance, so we read amyloid under no-$B$ as a broader cross-modal disease-state signal rather than a $P$-localized one. A patient-level bootstrap over the fixed test cohort placed the $P$-block amyloid and tau readouts above chance in every setting, with markedly wider intervals under no-$B$.

\subsection{Ablation Studies}
\label{sec:res_ablation}

\begin{table}[h]
\caption{RPNS allocation sensitivity. $R/P/N/S$ block widths, $d_z$ total; BAcc/macro-F1 \%, NLL raw. Mean$\pm$std over five seeds; bold marks best.}
\label{tab:dim}
\centering
\footnotesize
\setlength{\tabcolsep}{3.8pt}
\begin{tabular}{cccc c ccc}
\toprule
\textbf{R} & \textbf{P} & \textbf{N} & \textbf{S} & \textbf{$d_z$}
& \textbf{BAcc}$\uparrow$ & \textbf{macro-F1}$\uparrow$ & \textbf{NLL}$\downarrow$ \\
\midrule
 4 &  8 &  8 &  4 &  24 & $69.47{\pm}1.78$ & $70.13{\pm}1.74$ & $0.695{\pm}0.004$ \\
12 & 12 & 12 & 12 &  48 & $70.09{\pm}1.58$ & $70.68{\pm}1.42$ & $0.682{\pm}0.010$ \\
16 &  8 &  8 & 16 &  48 & $68.72{\pm}2.15$ & $69.57{\pm}2.06$ & $0.690{\pm}0.011$ \\
16 & 16 & 16 & 16 &  64 & $69.37{\pm}2.36$ & $70.39{\pm}2.16$ & $0.693{\pm}0.013$ \\
 8 & 32 & 16 &  8 &  64 & $70.62{\pm}1.74$ & $71.10{\pm}1.41$ & $0.684{\pm}0.012$ \\
32 & 32 & 32 & 32 & 128 & $69.28{\pm}2.17$ & $70.14{\pm}1.80$ & $0.697{\pm}0.013$ \\
\midrule
\textbf{8} & \textbf{16} & \textbf{16} & \textbf{8} & \textbf{48} 
& $\mathbf{71.87{\pm}0.89}$ & $\mathbf{71.92{\pm}0.62}$ & $\mathbf{0.681{\pm}0.006}$ \\
\bottomrule
\end{tabular}
\end{table}

\textbf{RPNS dimension allocation.} Table~\ref{tab:dim} checks whether the main result is explained by latent width alone. At the same total width $d_z=48$, the role-aware $8/16/16/8$ allocation outperforms both symmetric $12/12/12/12$ and reversed $16/8/8/16$ variants, and larger widths do not close the gap. We treat this as a sensitivity sweep, not a claim of unique optimality.

\begin{table}[h]
\caption{Component ablations. Metrics \%; Brier/NLL raw. Mean$\pm$std over five seeds.}
\label{tab:comp}
\centering
\footnotesize
\setlength{\tabcolsep}{4pt}
\resizebox{\columnwidth}{!}{%
\begin{tabular}{lcccc}
\toprule
\textbf{Configuration}
& \textbf{BAcc}$\uparrow$ & \textbf{AUROC}$\uparrow$
& \textbf{Brier}$\downarrow$ & \textbf{NLL}$\downarrow$ \\
\midrule
\textbf{Full CoPoE}      & $71.87{\pm}0.89$ & $85.80{\pm}0.33$ & $0.408{\pm}0.003$ & $0.681{\pm}0.006$ \\
\;w/o RPNS partition     & $68.10{\pm}2.24$ & $84.41{\pm}0.35$ & $0.421{\pm}0.004$ & $0.708{\pm}0.011$ \\
\;w/o residual pathway   & $68.83{\pm}1.71$ & $85.62{\pm}0.68$ & $0.434{\pm}0.009$ & $0.727{\pm}0.018$ \\
\;w/o block alignment    & $69.43{\pm}0.62$ & $84.28{\pm}0.49$ & $0.422{\pm}0.006$ & $0.707{\pm}0.015$ \\
\;w/o biomarker heads    & $72.32{\pm}1.17$ & $86.05{\pm}0.65$ & $0.410{\pm}0.004$ & $0.687{\pm}0.012$ \\
\;w/o RPNS \& residual   & $66.53{\pm}1.59$ & $82.77{\pm}0.87$ & $0.439{\pm}0.015$ & $0.744{\pm}0.032$ \\
\bottomrule
\end{tabular}%
}
\end{table}

\textbf{Model components.} Table~\ref{tab:comp} ablates the major components in the all-modality setting, reporting classification metrics alongside proper scores. The \emph{w/o RPNS partition} control is a capacity-matched generic-latent baseline: it replaces the structured latent with a single unstructured $48$-d latent, keeping the encoders, Gaussian PoE, KL, subset training, and residual MLP, while block-specific anchoring and cascade losses are disabled and the biomarker heads read the full latent. Its degradation in BAcc and AUROC supports the RPNS-structured fusion design over a flat Gaussian-PoE rule, though it does not isolate the partition geometry from the block-dependent losses. Removing the residual pathway mainly degrades BAcc, Brier, and NLL, while AUROC remains close to the full model ($85.62$ vs.\ $85.80$); thus, the residual branch improves thresholded classification and proper scoring rather than replacing the RPNS evidence accumulator; removing the biomarker heads leaves BAcc and AUROC essentially unchanged but slightly worsens Brier and NLL, so we retain the PET heads for probability quality and probing rather than to improve classification accuracy. Joint removal of the partition and residual branch gives the largest degradation, so the residual branch does not subsume the disease-coordinate posterior.

\subsection{Discussion and Limitations}
\label{sec:res_discussion}

The all-modality and observed-subset results support CoPoE as a disease-coordinate evidence accumulator whose advantage grows as complementary evidence is added to the same RPNS posterior. The flat Gaussian-PoE and residual-pathway ablations (Sec.~\ref{sec:res_ablation}) indicate that the gains are not explained by generic PoE fusion or by the residual branch alone; rather, the RPNS fusion layer provides the subset-invariant evidence accumulator, while the residual branch improves final classification and probability quality.

These conclusions should be interpreted within a standardized ADNI non-PET embedding benchmark. The common complete-case observed-subset panel controls subject composition across missing-modality settings, but it does not by itself establish deployment performance under natural acquisition policies or external cohort shifts. The RPNS axes should also be interpreted as disease-oriented coordinates: PET supervision directly tests the pathology block, whereas independent validation of the $R$, $N$, and $S$ blocks remains future work.

CoPoE further makes a modeling trade-off by using diagonal Gaussian experts, which preserves closed-form observed-expert fusion for arbitrary modality subsets but does not model cross-coordinate covariance. Larger external cohorts with natural missingness patterns and broader PET-labeled samples will be needed to assess site robustness, missingness mechanisms, and subgroup performance.

\section{Conclusion}
We presented CoPoE, a disease-coordinate masked Product-of-Experts framework for robust missing-modality AD diagnosis. CoPoE treats diagnosis from any observed modality subset as observed-evidence accumulation in RPNS coordinates: absent modalities add no factor in the RPNS path, and the posterior remains decomposable across subsets. On ADNI, CoPoE achieves the best all-modality performance and the highest mean AUROC across all $15$ observed-modality subset conditions among standardized adaptations of strong missing-modality baselines, including PRA-PoE, with substantially better ECE, Brier score, and NLL. PET readouts further show $P$-block enrichment under full evidence, with tau-related signal retained without biospecimen input. Extensive evaluations demonstrate that missing-modality diagnosis can be framed around measured-evidence accumulation in a decomposable disease-oriented coordinate system, without synthesizing unobserved inputs in the RPNS path. Future work should validate CoPoE on external cohorts, model missingness mechanisms more explicitly, and assess subgroup robustness.

\bibliographystyle{ieeetr}
{\let\oldbibitem\bibitem
\renewcommand{\bibitem}{\setlength{\itemsep}{-0.10ex}\oldbibitem}
\bibliography{refs}}

\end{document}